# A Two Stage Classification Approach for Handwritten Devanagari Characters

Sandhya Arora
*Meghnad Saha Institute of Tech.,WBUT Kolkata*
sandhyabhagat
@yahoo.com

Debotosh Bhattacharjee
*Jadavpur University Kolkata*
debotoshb
@hotmail.com

Mita Nasipuri
*Jadavpur University Kolkata*
mitanasipuri
@yahoo.com

Latesh Malik
*Nagpur University Nagpur*
lateshmalik
@rediffmail.com

*Abstract*

*The paper presents a two stage classification approach for handwritten devanagari characters The first stage is using structural properties like shirorekha, spine in character and second stage exploits some intersection features of characters which are fed to a feedforward neural network. Simple histogram based method does not work for finding shirorekha, vertical bar (Spine) in handwritten devnagari characters. So we designed a differential distance based technique to find a near straight line for shirorekha and spine. This approach has been tested for 50000 samples and we got 89.12% success.*

## 1. Introduction

OCR work on printed Devnagari script started in early 1970s. Among the earlier pieces of work, some of the efforts on Devnagari character recognition are due to Sinha [1,7,8] and Mahabala [1]. Sethi and Chatterjee [5] also have done some earlier studies on Devnagari script and presented a Devnagari hand-printed numeral recognition system based on binary decision tree classifier. They [6] also used a similar technique for constrained hand-printed Devnagari character recognition. They did not show results of scanning on real document pages. The first complete OCR system development of printed Devnagari is perhaps due to Palit and Chaudhuri [4] as well as Pal and Chaudhuri [3]. For the purpose some standard techniques have been used and some new ones have been proposed by them. The method proposed by Pal and Chaudhuri gives about 96% accuracy. Recently, a system for hand-written numeral recognition of Devnagari character is proposed [2]. A few of these work deals with handwritten characters of Devnagari. Because of the complexities involved with devanagari script, already existing methods can not be applied directly with this script.

We used differential distance based technique for identifying the shirorekha and spine, as the simple histogram based technique does not identify the straight line for handwritten devnagari character. Priority based search technique is used to identify the valid movement for shirorekha. A priority mask is moved from rightmost pixel of character, as the movement can go only in directions shown in mask for shirorekha. After identifying shirorekha and spine, preliminary classification is done based on these features. For various grouping of characters intersection features are calculated and feature vector is fed to a backpropagation feedforward neural network.

## 2. Specific features of Devnagari script

Devnagari script has about 11 vowels ('svar') and 33 consonants or ('vyanjan'), 10 numerals along with modifier symbols. Script has its own specified composition rules for combining vowels, consonants and modifiers. Vowels are used to produce their own sound or they are used to modify the sound of a consonant by attaching an appropriate modifier in an appropriate manner with them. Modifier symbols are placed on top, bottom, left, right or on a combination of these. The consonants may also have a half form or shadow form. A half character is written touching the following character resulting in a composite character. In part, Devanagari owes its complexity to its rich set of conjuncts.

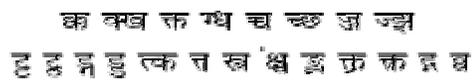
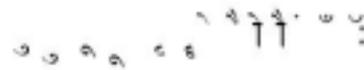

Figure 1. Half and shadow characters          Figure 2. Modifiers

Devanagari characters are joined by a horizontal bar (Shirorekha) that creates an imaginary line by which Devanagari text is suspended. A single or double vertical line called a Danda (Spine) was traditionally used to indicate the end of phrase or sentence.

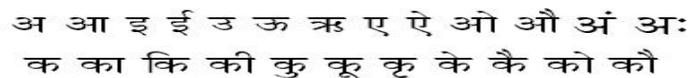

Figure 3. Non-compound devnagari characters

Above examples give an idea of various ways in which a character takes a form of composite character. These lead to a large number of character fusion and character overlaps either due to shape or due to modifiers. This makes it very difficult to develop and OCR for devnagari script.

## 3. The Recognition process

The process of conversion of scanned image into a text document primarily consists of the following steps:

### 3.1. Preprocessing

The preprocessing steps removes any distortions or discontinuity in the input character and convert the characters into a form recognizable by the detection procedure. It consists of following steps:

**3.1.1. Size Determination** : This step determines the approximate dimension of the character by forming a tight fit rectangular boundary around the character.

**3.1.2. Distortion Removal :** We use thickening, thinning and pruning for removing distortions [13]. The image is thickened first and then thinned to convergence. This gives us a smooth one-pixel wide image [9] of the character, which is pruned to remove the small projections resulting from the thinning algorithm. Small characters should be distortion-free.

**3.1.3. Normalization** : After thinning character is scaled to 100 X 100 pixels using affine transformation..

### 3.2. Shirorekha Detection

We use the shirorekha in preliminary grouping of the characters. It is assumed that the first pixel available for a given character, when looked from right to left, is the shirorekha, i.e nothing can go past the shirorekha as is normal practice. This last pixel is traced out for the character and the shirorekha movement is detected by priority based neighborhood search (Fig-4). Differential distance based technique (Fig-6) on the detected shirorekha determines whether it is a near straight line or not. This technique is based on calculating the successive differences of the distances of upper envelop of the curve from the reference line. Reference line is top of the character image. Also, the shirorekha should terminate in an open end. If not, the shirorekha is rejected and the character is said to have no shirorekha

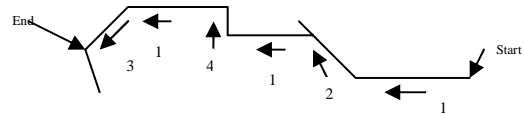

Figure 4. The priority mask                Figure 5. Working of priority mask

### 3.3. Spine Detection

The spine detected as the downward near straight line looking from right of each character. Again, near straight lines are detected by differential distance based technique (Fig-7). A straight line to qualify as a spine should be at least $3/4^{th}$ the height of the character (an assumption is made that the lower matra doesn't occupy more than 1/4th of the character height). If no such near straight line exists, the character is considered without a spine. A spine can only exist if a shirorekha exists. If two such spines are found, the first one from right is considered to be the matra. There cannot be more than two spines.

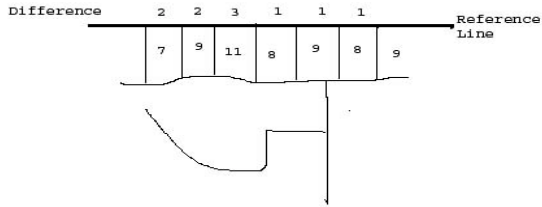 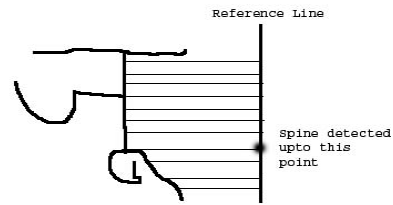

Figure 6.  Detecting shirorekha                Figure 7. Detecting spine

### 3.4. Intersection Features

Feature is a point of human interest in an image, a place where something happens. It could be an intersection between two lines, or it could be a corner open end or it could be just a dot surrounded by space. These relationships are used for character identification. Intersection features are unique for different characters, hence the feature points are exploited for the task of character recognition.

Each handwritten character can be adequately represented within 16 segments (each of size 25 X 25 pixels) and hence 32 features for each character can be used as input to neural network. We are using a discrete structural approach and breaking the character boundary into 16 segments. For each segment number of intersection points, number of open ends is being calculated.

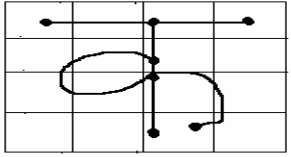

**Figure 8. Intersection points of character**

## 4. Classification

Before applying Neural network, a prelimimary classification is performed for better results. The presence and position of spine divides character set of devanagri into different classes, so the entire character map for Devanagari characters (excluding matras) can be grouped according to the following criteria

### 4.1. *Shirorekha Continuity*

Some characters contain a *shirorekha* throughout, while the others contain a partial shirorekha or no shirorekha. Thus three groups can be obtained by this method.

### 4.2. *Spine Location*

Another important aspect of the character is its "spine". Characters can be divided into three groups i)End Spine (The spine is the rightmost part of the character ii)Mid Spine (The spine exists in between, i.e. there are some parts of the character on either side of the spine iii)No Spine (There are no spines in these characters). By taking an intersection of the above two properties, the entire character map can be divided into small groups. This eases the task of recognition.

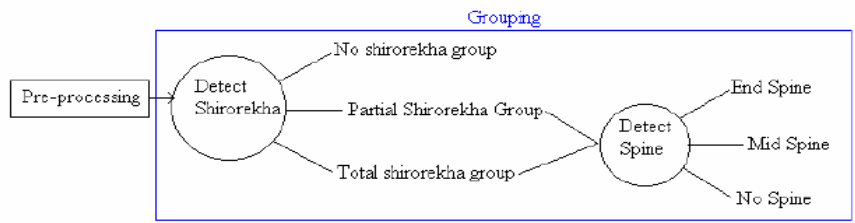

**Figure 9. Preliminary classification**

### 4.3. *Neural Network Architecture*

For each group characters, feature vector of size 32, consisting of intersection points and end points value in each segment is calculated and NN is designed using this input. A 3-layer feed forward architecture with back propagation is considered with one input layer of 32 neurons, one hidden layer, and one output layer. We tested for different hidden layer neurons but this maximum accuracy is achieved by 40 neurons in hidden layer. The Scaled conjugate gradient training method is applied. The scg training parameters of interest are: learning rate 0.01 momentum factor 0.95 and minimum gradient .00000001. The number of hidden layer neurons has been changed to get best results.

## 5. Experimental Result

These experimental results are obtained for 50000 data samples. Out of which 35000 samples are used for training and 15000 for testing.

Table 1. Results

|  | Test data accuracy | Training data accuracy | Some Example data samples |
|---|---|---|---|
| Partial shirorekha, End spine group characters | 100% | 100% | ध भ भ |
| Total shirorekha, End spine group characters | 89.12% | 98.3% | च ख ब |
| Total shirorekha, Mid spine group characters | 90% | 99.43% | फ क |
| Total shirorekha, No spine group characters | 95% | 98.67% | ङ ह ट ठ |

## 6. Conclusion

Development of handwritten Devnagari OCR is still a challenging task in Pattern recognition area. Detecting the near straight line in handwritten character for detecting shirorekha and spine can not be done using conventional methods. We purpose a differential distance based technique and priority based search mechanism for this. We studied the intersection feature which is giving 89.12% accuracy.